# Online and Scalable Model Selection with Multi-Armed Bandits

Jiayi Xie, Michael Tashman, John Hoffman, Lee Winikor, Rouzbeh Gerami

Copilot AI

Xaxis

New York, USA

{jiayi.xie, michael.tashman, john.hoffman, lee.winikor, rouzbeh.gerami}@xaxis.com

## ABSTRACT

Many online applications running on live traffic are powered by machine learning models, for which training, validation, and hyper-parameter tuning are conducted on historical data. However, it is common for models demonstrating strong performance in offline analysis to yield poorer performance when deployed online. This problem is a consequence of the difficulty of training on historical data in non-stationary environments. Moreover, the machine learning metrics used for model selection may not sufficiently correlate with real-world business metrics used to determine the success of the applications being tested. These problems are particularly prominent in the Real-Time Bidding (RTB) domain, in which ML models power bidding strategies, and a change in models will likely affect performance of the advertising campaigns. In this work, we present Automatic Model Selector (AMS), a system for scalable online selection of RTB bidding strategies based on real-world performance metrics. AMS employs Multi-Armed Bandits (MAB) to near-simultaneously run and evaluate multiple models against live traffic, allocating the most traffic to the best-performing models while decreasing traffic to those with poorer online performance, thereby minimizing the impact of inferior models on overall campaign performance. The reliance on offline data is avoided, instead making model selections on a case-by-case basis according to actionable business goals. AMS allows new models to be safely introduced into live campaigns as soon as they are developed, minimizing the risk to overall performance. In live-traffic tests on multiple ad campaigns, the AMS system proved highly effective at improving ad campaign performance.

## Keywords

Display advertising; multi-armed bandit; model selection

## 1. INTRODUCTION

When widely deploying a new machine learning model in a live-traffic environment, it is necessary to assess the performance of the model beforehand, particularly according to the metrics directly tied to success of the application in the real world. It is commonly observed that models that perform well in offline tests may not fare similarly once facing real traffic, particularly in non-stationary environments. The ideal method of testing a model is therefore to conduct a controlled real-world experiment, whenever possible.

Real-world tests present challenges as well. When conducting an A/B test, typically an equal budget must be allocated to each branch of the test. However, equally allocating resources to each strategy ignores the fact that there may be enough data before the end of an A/B test to adequately judge their performance. This results in more resources being allocated to poorly performing strategies than necessary. Furthermore, a good deal of friction and additional cost exists in arranging online tests, and in many cases these tests require close attention, so that models that perform particularly poorly could be stopped early, while those that perform well may be widely deployed without unnecessary delay.

These issues are typical in the domain of Real-Time Bidding (RTB), a common method of allocating online advertising inventory. RTB utilizes auctions for individual units of inventory for which bidding methodologies must be set in advance. These methodologies can be simple—such as a single flat bid amount for any ad on a certain website—but simple methods are highly inefficient and increasingly uncommon. Instead, a typical approach uses historical data to train machine learning models to predict click or conversion probabilities for an ad impression opportunity. This probability is the basis for calculating the bids for the auction [Perlich et al., 2012].

Building such models requires a large amount of data for training, testing, and tuning hyperparameters. It further requires a set of machine learning metrics to assess the models before deployment, such as f-measure, AUC, or cross-entropy. Even when this is done, it is common for models trained and validated in this way, when deployed online, to not perform as well as the offline tests imply.

There are several reasons for this. The historical training and test data contain a set of biases. For example, they may include only the data from that auctions that were won, and lack information from the lost ones. Furthermore, the datasets are of limited size and may not offer a proper view of the entire input space, and many feature combinations may not have been observed with sufficient frequency. RTB markets are also volatile and the distribution in the feature space can be non-stationary.

The particular metrics being used must also be considered. Advertising campaigns are typically judged through real-world business analytics, including custom-designed metrics, which are very likely different from the ML metrics used in the standard offline model-selection methods.

Modeling methodologies may also perform differently when deployed to different advertisers and market situations. For example, simpler models may perform better for campaigns with historically low budgets and relatively smaller training data, while more complex models may be needed for larger campaigns, where larger data sets are available.

Finally, the standard model selection process can be time consuming and costly, requiring humans in the loop for individual modeling candidates, to go through offline and online tests. The online tests are likely to be set up and monitored separately, and the decisions regarding them made on a case-by-case basis, making the process non-scalable and prone to inefficiencies.

To address these drawbacks, we developed Automatic Model Selector (AMS), a platform using Multi-Armed Bandits to manage the model selection process in an online, safe and scalable fashion, based on real-world metrics. Different arms correspond to different ML models to be activated within a bidding strategy. Normally, one arm is assigned to the current production model, while other arms correspond to alternative models, where modeling methodology, training data, or features of the input space may be different.

Every time an arm is selected, its corresponding ML model is activated to power a live bidding strategy, remaining in effect for a fixed period of time until a different arm is chosen. The Multi-Armed Bandit algorithm then swaps different models sequentially within the same bidding strategy, providing them the opportunity to be tested against live traffic, and gradually settling on the model that performs best according to the key performance indicator (KPI) set for the campaign. MAB assures a well-balanced exploration-exploitation process in which the best model can be picked with a relatively low cost of exploration.

AMS performs the model selection process using online data, and thus avoids biases contained in historical data. It also makes selections based on relevant business metrics, rather than the standard ML metrics commonly used to evaluate models. This selection process is carried out in an automatic, scalable and repeatable way. The same set of arms can be activated on separate bidding strategies, allowing different choices to be made based on specifics of the campaign and its attributes. Once a modeling choice is finalized, we can expect a smooth transition to deployment to production, avoiding unexpected performance fluctuations caused by the model switch.

## 2. BACKGROUND
In this section, we will discuss previous work in multi-armed bandit model selection, and some mathematical and algorithmic choices in approaching the multi-armed bandit problem.

### 2.1 Prior Work
A multi-armed bandit approach to selecting models can be applied to many domains. In principle, there are several main requirements: that it is possible for multiple models to be activated sequentially; that it is not conclusive from offline testing as to which model would be most performant; that all models can be evaluated on the same criteria; and that models can return results with a short feedback loop, in sufficient volume to attain statistical significance.

In Shukla et al. [2019], an MAB approach is used to choose the best pricing model for airline ancillary goods. Authors reported that multiple models were developed which tended to outperform human choices on average—but no single model was consistently better than the others in all contexts. The authors built a multi-armed bandit approach using Thompson sampling, and this method led to a significant increase in revenue and total number of conversions compared to a randomly selected model. Zeng et al. [2016] used an MAB approach to build a recommender system, with the explicit assumption that the underlying environment is non-stationary, so that the reward function is expected to drift as a function of time. In Felício et al. [2017], authors use MAB to handle the cold start problem in recommender systems—in cases in which there is no information about a user, the MAB is used to select between various recommendation models.

### 2.2 Multi-Armed Bandit Algorithms
Various algorithms exist to solve the MAB problem. Thompson sampling is a popular approach, which requires deriving a posterior distribution over the KPI of interest for each arm, given the observation data and a prior—then selecting each arm with the probability that it attains the highest performance [Agrawal & Goyal 2013]. However, it may be practically difficult to derive a posterior and the best approach for assigning such a probability distribution is not always obvious.

An effective alternative is softmax—also known as Boltzmann exploration—in which a *temperature* hyperparameter is tuned to balance exploration and exploitation [Chapelle & Li 2011, Tang et al., 2013]. Upper confidence bound (UCB) [Kealbling, 1993] takes each arm's reward uncertainty into account and chooses the arm with the highest optimistic reward.

Based on the suggestions from [Mäkinen, 2017], we have chosen to adopt a "decay ε-greedy" approach due to its simplicity and relatively good performance [Auer et al., 2002]. ε-greedy requires setting a value for $\varepsilon$, a parameter representing the strength of the exploration: at each step the algorithm selects best arm with probability $1 - \varepsilon$, and with probability $\varepsilon$ it will choose to explore, selecting a model uniformly at random from the set of all candidate models (including the best performing model). That is, for a given value of $\varepsilon$, the activation probability for the best performing and alternative arms are given by

$$P_{best} = (1 - \varepsilon) + \varepsilon/M$$
$$P_{other} = \varepsilon/M$$

where $M$ is size of the candidate set.

In decay ε-greedy ε is lowered over time, so that selection gradually becomes greedier, allowing the arm with the best empirical performance to eventually be the only choice. Here we choose the following functional form for $\varepsilon(t)$:

$$\varepsilon(t) = \varepsilon_0 \max(0, (1 - t/\alpha))$$

Where $t$ is time, and $\alpha$ is a scaling parameter controlling the speed of decay.

## 3. IMPLEMENTATION
This section will discuss the implementation of our automatic model selector in practice. The AMS system is designed to be *general*, *live*, and *safe*. In designing the system to be *general*, it is intended to act as a flexible mechanism for optimizing any model used on live RTB traffic, and able to base its decisions on any performance measure. (We typically use this system with click or conversion models, which are measured by KPIs such as CPC, CPA or CTR.) In working *live*, we avoid the drawbacks and biases of offline testing, and the possibility of relying on old or stale data. AMS ensures *safety* through its framework of minimizing regret. Because it is able to learn which models over- and underperform, AMS efficiently selects between them to maximize performance wherever possible. These decisions are always based on the most recent available performance information, enabling AMS to safely run on live, active campaigns. This offers significant advantages over A/B testing which allocates far more time and spend to underperforming models. This creates an environment in which it is safe to deploy fresh models, despite uncertainty as to the models' expected performance.

Here we discuss the specifics of our implementation in the context of our typical use case: optimizing click or conversion models on live RTB traffic for a particular campaign, with the goal of optimizing a single KPI. This section contains an overview of the individual components in the AMS system, and an explanation of how activation probabilities are calculated.

## 3.1 Overview

The AMS system has three major components (Figure 1):

(1) **ML Model Trainer** – handles training for all candidate models, typically CTR prediction models used in RTB bidding strategies, and usually expressed as either logistic regression or factorization machines. Each day, the model trainer collects all log-level data from the campaign and retrains each candidate model.

(2) **Performance Monitor** – logs the performance of each model against live traffic. Key performance indicators such as CPC, CPA, and CTR, and other custom indicators are calculated. These KPIs are computed individually for each of the models, typically once a day, using only the impressions purchased when that model was active within the bidding strategy.

(3) **MAB Model Selector** – typically runs every 15 minutes, using MAB to choose a new model to be activated as the core of the bidding strategy. We keep the run time for each step relatively short so that the models generally get to be tested against similar market environments.

## 3.2 Multi-Armed Bandit Algorithm

In this context each arm corresponds to a model trained by the model trainer. The models are selected sequentially by the MAB Model Selector to power the bidding strategy running on the campaign. We have chosen a decay ε-greedy algorithm for the model selection step. Despite its relative simplicity, it works effectively for our use case.

At each time interval the model with the best overall performance KPI is given a probability of 1-$\varepsilon$ to be activated. Underperforming models are given a non-zero probability of being selected, since performance of the models may change over time. We lower ε over time as performance estimates for the models stabilize and exploration becomes less important. Given the short feedback loop, this performance data quickly becomes available following deployment of a model.

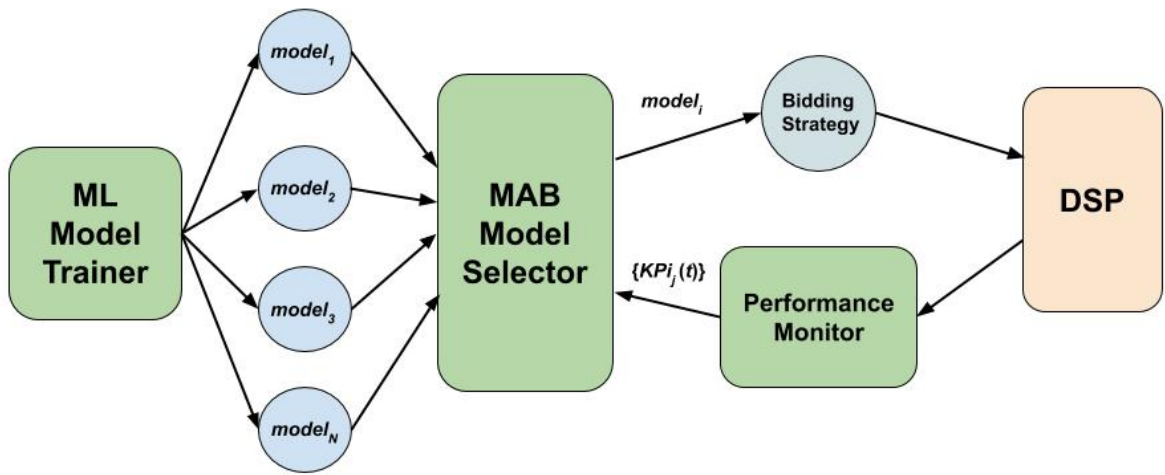

**Figure 1.** Components for the AMS system. ML Model Trainer provides trained ML models as the arms available to the MAB algorithm, run by MAB Model Selector. The selected model powers a bidding algorithm for a live RTB campaign running on a DSP. The campaign's performance KPIs are tracked by the Performance Monitor, based on which the selection probabilities for the arms are updated. Models are swapped every 15 mins and performance KPIs are updated once a day.

## 4. EXPERIMENTAL RESULTS

In this section, we cover experimental results from running AMS in two different test cases. The design of the MAB arms varies between tests; but in both cases, there is uncertainty as to the models' expected performance, AMS makes it safe to evaluate and deploy fresh models online.

### 4.1 Role of Size of Training Data

In machine learning, additional training data is genrally expected to improve model performance. In case of RTB campaigns, however, given the volatility of bidding environments, older training data may not be a good representation of the current market conditions and including them in the training set could reduce model performance. It is also unclear if the shelf life of training data is the same across campaigns, as some campaigns target markets that are more stable than others. Since a single answer is unlikely to fit all campaigns, this problem is a good use case for AMS.

There are two arms introduced in this test. Both are logistic regression models for CTR prediction, but the arms are trained with different lookback window sizes for the historical data. One model is trained with a seven day lookback window, while the other is trained with 60 days. (*Model7* and *model60*, respectively.) Models are trained using all data from the parent campaign, which contains multiple sub-campaigns.

The results are plotted in Figure 2, where we plot test AUC as an offline evaluation measure, as well as the cumulative CTR (*cumCTR*), with a lookback window of 30 days, measured against online traffic. *model7* initially has the best AUC score; however, *model60* begins to get the highest AUC score around day 7 and continues to do so for the remainder of the test. The online test has a similar, but not identical, overall trend.

Several points are noteworthy – (1) more training data seems to result in a better-performing model, as judged by both offline and online metrics. (2) the offline and online metrics generally maintain a consistent relationship, but notable exception is at the beginning phase of the campaign where higher AUC does not translate to higher live CTR. (3) AMS is effective in picking the better performing model, while leaving room for exploration with the extent of exploration decreasing over time.

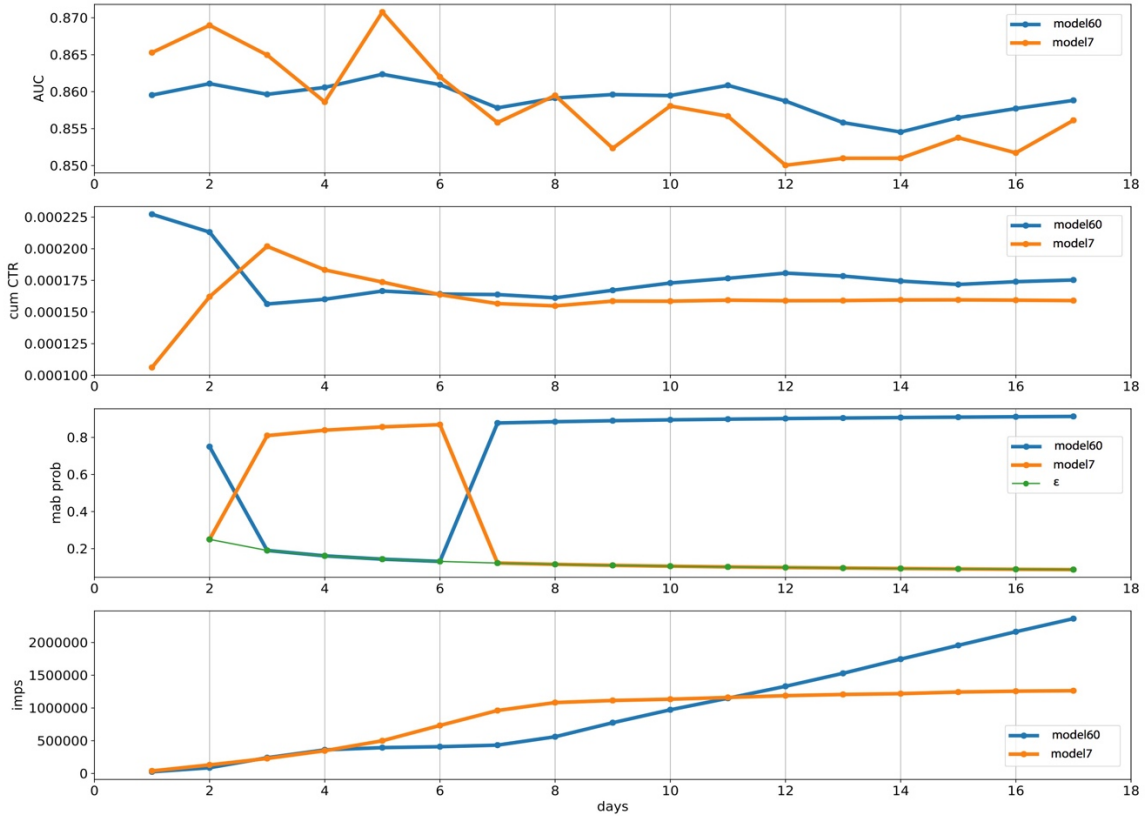

**Figure 2**. Results for the test described in section 4.1. From top to bottom: AUC: an offline metric to measure the performance of a model created on that day, is calculated using the offline test data set. Cumulative CTR: an online metric that measures the performance of each model based on real traffic, which has a 30-day lookback window. MAB probability: the activation probability of each model ("arm"). Cumulative impressions: the number of impressions assigned to each model.

*4.2 Inclusion of additional features*

In this test, the two arms correspond to logistic regression models with different numbers of features. In the *modelControl*, we use all the data from parent-campaign as a whole for modeling, without differentiating sub-campaigns. In the *modelTest*, we cross the sub-campaign ids with original features, to help the model learn to differentiate between sub-campaigns. For example, feature like $domain_A$ becomes $id_1 \diamond domain_A$, $id_2 \diamond domain_A$… $id_N \diamond domain_A$, where *N* equals to the number of sub-campaigns within the parent campaign. As a result, the number of features expands N times in the test group. For our test, *modelControl* has about 1000 features, and *modelTest* has about 7000 features. While additional features may improve the model, we also increase the risk of overfitting. The AMS results are shown in Figure 3.

Judging by test AUC (the offline metric), *modelControl* generally yields slightly better performance than *modelTest*. However, in the online test *modelControl* significantly outperforms *modelTest*. As a consequence, higher MAB probabilities and more impression traffic are gradually assigned to *modelControl*.

This demonstrates a case where offline testing metrics do not fully align with online KPI performance. The AMS system effectively detected the arm with the best performance on live traffic, while minimizing wasted resources by gradually moving spend away from the poorly performing option.

## 5. CONCLUSION

In this paper, we presented a multi-armed bandit method for optimizing real-time model selection for the RTB campaigns, with the goal of improving campaign KPIs, such as CTR. The MAB approach allows for low-friction deployment of new predictive models against real traffic, adjusting the amount of traffic given to each model as a consequence of its performance. MAB therefore makes it possible to avoid excessive offline testing, and offers an efficient alternative to costly A/B tests by quickly detecting poor model performance and directing spend away from them. This method was demonstrated to be effective in online testing.

For further research, we would propose improving the way we monitor model or KPI performance. For example, when the reward distribution is non-stationary, various approaches can be applied to address the temporal uncertainty [Besbe et al., 2014]. We would also propose addressing the challenge of applying AMS in an environment with delayed feedback [Chapelle, 2014], for example, campaigns that optimize towards conversions in which the conversion takes place long after the impression was bought. Such an approach would require an inference into the current state of the system, despite only having access to older information, and would likely involve a module such as a Smith predictor [Sourdille, et al., 2003].

## ACKNOWLEDGMENTS
The authors would like to thank Victor Seet, Jacob Wan, Tobias Sutters, Bhuvan Allanki, Ben Yi, and Jwalin Shah for their contributions.

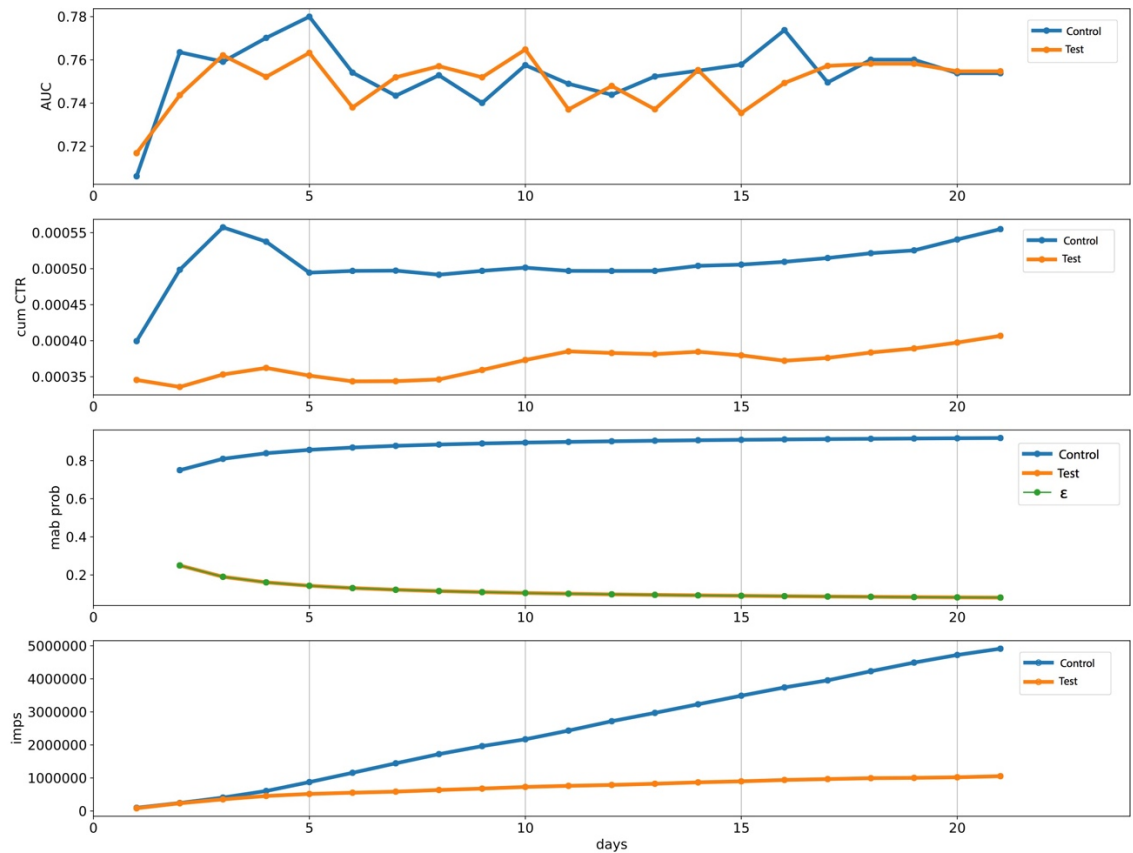

**Figure 3.** Results for the test described in section 4.2.